\documentclass[runningheads]{llncs}

\usepackage{subfig}
\usepackage{graphicx}
\usepackage{amsfonts}
\usepackage{eucal}
\usepackage{amsmath}
\usepackage{booktabs}
\usepackage{lipsum}
\usepackage[algo2e]{algorithm2e}
\usepackage{todonotes}
\usepackage{xcolor}
\usepackage{comment}
\usepackage{multicol}
\usepackage{multirow}
\usepackage{tabularx}
\usepackage{fixltx2e}
\usepackage{float}
\usepackage{algorithm}
\usepackage{algorithmicx}
\usepackage{algpseudocode}
\usepackage[T1]{fontenc}
\usepackage{tipa}

\begin{document}
\titlerunning{Interpretable Early Detection of PD through Speech Analysis}
\title{Interpretable Early Detection of Parkinson's Disease through Speech Analysis}
%
%
\author{Lorenzo Simone \inst{1,2}\orcidID{0000-0002-5010-7733} \and Mauro Giuseppe Camporeale \inst{3}\orcidID{0000-0002-0323-5505} \and Vito Marco Rubino \inst{3} \and Vincenzo Gervasi \inst{1}\orcidID{0000-0002-8567-9328} \and Giovanni Dimauro\inst{3}\orcidID{0000-0002-4120-5876}}
\authorrunning{L. Simone et al.}
%
\institute{
\textsuperscript{1} Department of Computer Science, University of Pisa, Pisa, Italy \\
\textsuperscript{2} School of Public Health, Yale University, U.S.A. \\
\textsuperscript{3} Department of Computer Science, University of Bari, Bari, Italy
}

\maketitle              
\begin{abstract} Parkinson’s disease is a progressive neurodegenerative disorder affecting motor and non-motor functions, with speech impairments among its earliest symptoms. Speech impairments offer a valuable diagnostic opportunity, with machine learning advances providing promising tools for timely detection. In this research, we propose a deep learning approach for early Parkinson’s disease detection from speech recordings, which also highlights the vocal segments driving predictions to enhance interpretability. This approach seeks to associate predictive speech patterns with articulatory features, providing a basis for interpreting underlying neuromuscular impairments. We evaluated our approach using the Italian Parkinson's Voice and Speech Database, containing 831 audio recordings from 65 participants, including both healthy individuals and patients. Our approach showed competitive classification performance compared to state-of-the-art methods, while providing enhanced interpretability by identifying key speech features influencing predictions.

\keywords{Parkinson's disease \and deep learning \and explainable detection \and speech processing \and signal analysis}
\end{abstract}

\section{Introduction}
\label{sec:intro}
Parkinson's disease (PD) is a neurodegenerative disorder that progressively affects the central nervous system, characterized by the selective deterioration of neuronal cells and a subsequent decline in dopamine levels \cite{parkinson-new}. This degeneration impairs motor coordination, which in turn affects respiration, phonation, articulation, and prosody, leading to the hallmark physiological and anatomical changes seen in PD \cite{parkinson-coordination}. Speech difficulties are among the earliest signs, with patients often presenting with a soft voice, monotony, hoarseness, and imprecise articulation. Early diagnosis is crucial for effective intervention, and machine learning (ML) has shown strong potential in identifying speech patterns that serve as reliable disease indicators.

In previous literature, ML techniques have been extensively applied for PD detection using speech data. Traditional approaches rely on extracting handcrafted features from voice recordings, followed by classification using models such as Support Vector Machines (SVM), Random Forest (RF), K-Nearest Neighbors (KNN), and Logistic Regression \cite{rw1}. Other studies have incorporated feature selection techniques like Recursive Feature Elimination and feature importance ranking to refine predictive accuracy \cite{rw3}. Hybrid methods combining neural networks with Lagrangian SVM have also been explored to optimize classification performance \cite{rw2}. However, these approaches predominantly rely on static, precomputed features, limiting their ability to capture temporal dependencies in speech data.

While deep learning has been applied to PD detection, existing studies primarily focus on non-speech biomarkers such as rapid eye movement, olfactory loss, and cerebrospinal fluid markers \cite{rw4}. The current gap in temporal deep learning methods for modeling speech dynamics restricts the ability to apply explainability techniques, which could provide critical insights into the relationship between speech patterns and neuromuscular impairments, ultimately limiting diagnostic potential.

\section{Methodology}

A key challenge in speech analysis for PD prediction is the variability in language, recording equipment, audio quality, and environmental conditions, as well as individual differences among speakers. To better analyze the contribution of each pronounced word to vocal articulation, we segmented long speech recordings into smaller, consecutive units suitable for processing by a temporal neural network. An initial approach based on silence thresholds proved unreliable due to variations in audio intensity, leading to inconsistent segment lengths. To improve precision, we utilized word-level timestamps from an open speech recognition library, defining each segment as a fixed set of consecutive words. While this method offered better consistency, occasional timing inaccuracies caused partial word loss, particularly in fast speech. To mitigate this, we developed a hybrid approach combining word boundaries with amplitude-based cutting, ensuring robust segmentation for deep neural network processing.
\\
\noindent \textbf{Temporal Network for Speech Analysis.} The deep learning architecture is specifically designed to process audio signals segmented into speech word chunks, a crucial aspect of our application. Unlike traditional speech processing models, which typically operate on continuous audio streams, our approach handles discrete word chunks, allowing the model to focus on individual units of speech for more precise feature extraction and classification. 

Formally, each convolutional block is composed of a 1D convolutional layer, a batch normalization (BN) layer, and a ReLU activation. A convolutional layer at depth $m$ processes the input signal $\mathbf{X}^{(m-1)}$ and produces the output feature map $\mathbf{A}_k^{(m)}$, calculated as:

{\footnotesize
\[
\mathbf{A}_k^{(m)} = \mathnormal{ReLU}(\mathnormal{BN}(\sum_{i=1}^K \mathbf{W}_{ki}^{(m)} * \mathbf{A}_i^{(m-1)} + b^{(m)}_k)),
\]
}

where $\mathbf{W}_{ki}^{(m)}$ and $b^{(m)}_k$ are the weight and bias matrices, and $*$ represents the convolution operation. In our architecture, the first convolutional layer processes the word chunk input, generating 48 output channels with a kernel size of 3 and padding of 1. The second layer then takes 48 input channels and generates 96 output channels, also with a kernel size of 3 and padding of 1. The output from the convolutional blocks is flattened and processed through a fully connected layer.

\noindent \textbf{Interpretability analysis.} To interpret the contributions of different speech segments toward the prediction of Parkinson’s disease, we employed Gradient-weighted Class Activation Mapping (Grad-CAM) \cite{gradcam,ECGinterpretation}, adapted to our temporal speech signal processing scenario. Given a processed speech signal segmented in word chunks, we computed gradient class activation mappings by averaging the samples within respective classes. This allowed us to visualize the speech segments that positively or negatively impact the classification outcome. While preserving the original conceptual framework, we adapted Grad-CAM for one-dimensional temporal signals by analyzing the gradient of the final convolutional layer’s output with respect to the class score.

Formally, let $L(c)$ denote the final convolutional layer in our CNN model, and let $A_k^L(t)$ represent the activation map for the $k$-th feature map in this layer at temporal position $t$. The importance weights are determined by computing the global average pooling of the gradient of the output class score $y^c$ with respect to the feature maps $A_k^L$:

\begin{equation}
\alpha_k^c = \frac{1}{Z} \sum_t \frac{\partial y^c}{\partial A_k^L(t)},
\end{equation}

\noindent where $y^c$ is the output score for class $c$, $A_k^L(t)$ is the feature map at time $t$ for the $k$-th layer, and $Z$ is a normalization factor. We then compute a weighted combination of the activation maps:

\begin{equation}
L^c(t) = \sum_k \alpha_k^c A_k^L(t),
\end{equation}

\noindent This weighted sum provides insight into the specific portions of the input that are most influential in the prediction, as each feature map $A_k^L(t)$ is scaled by its corresponding importance weight $\alpha_k^c$. The resulting map $L^c(t)$ highlights the regions of the input that contribute significantly to the class prediction for $c$.

\section{Results and Discussion}
\label{sec:results}

The dataset used in this study is the "Italian Parkinson’s Voice and Speech Database" \cite{parkinson-dataset}, containing 831 audio files from 65 individuals (37 healthy, 28 with Parkinson’s disease) aged 19-80. Participants recorded various samples, including phonetically balanced texts, vowel pronunciations, and syllables ("pa" and "ta"). The results presented in Table \ref{tab:metrics}, obtained through a rigorous repeated stratified holdout procedure conducted across nine iterations, demonstrate that our approach performs competitively with simpler models such as KNN, SVM, RF, and Gradient Boosting (GB) on speech signals. Our model consistently achieves higher performance than baseline classifiers across all metrics, with statistically significant improvements in accuracy, recall, and F1-score ($p < 0.001$), and marginal significance in precision ($p = 0.061$). Beyond performance, the model highlights the vocal features most influential in each prediction, offering both interpretability and potential relevance to support clinical insights.

\begin{figure}[t]
    \centering
    \begin{minipage}{0.48\textwidth}
        \centering
        \includegraphics[width=\textwidth]{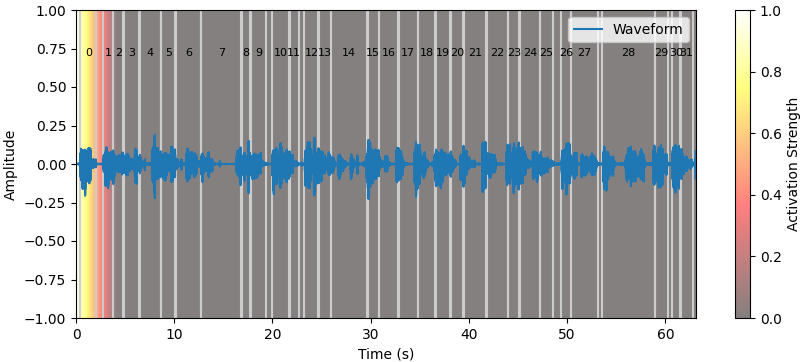}
        \label{fig:PD}
    \end{minipage}\hfill
    \begin{minipage}{0.48\textwidth}
        \centering
        \includegraphics[width=\textwidth]{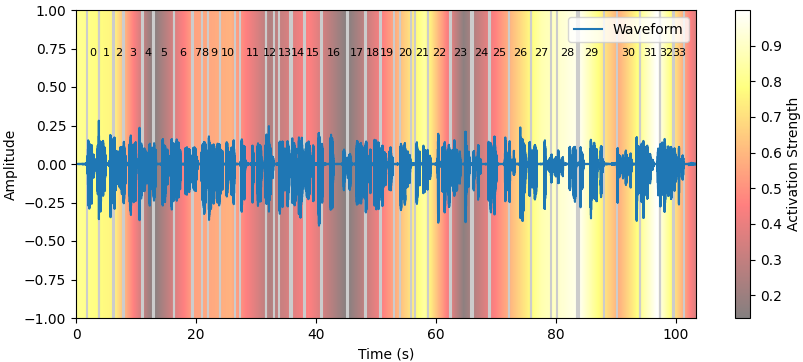}
        \label{fig:EHC}
    \end{minipage}
    \caption{Activation contributions for each audio segment, depicting differences between healthy (left) and Parkinson's disease (right) speech patterns.}
    \label{fig:comparison}
\end{figure}

\footnotetext[1]{Extended speech signals refer to two readings of a phonemically balanced text, each containing an average of 124 words, with a 30-second pause between them.}

The interpretability framework’s heatmaps (Figure \ref{fig:comparison}) enabled us to isolate the key speech segments contributing to Parkinson’s disease detection in our analysis. For each recording, segment activation values were normalized, and the 90\textsuperscript{th} percentile was computed to establish a threshold for significance. Segments exceeding this threshold were selected as the most informative (Table \ref{tab:pd}). Analysis of speech recordings from patients revealed a higher prevalence of front vowels (/\textit{i}/, /\textit{e}/, /\textit{y}/), plosive alveolar consonants (/\textit{t}/, /\textit{d}/), and fricatives (/\textit{s}/, /\textit{z}/), which require precise tongue and lip coordination. These articulatory demands pose challenges for individuals with Parkinson’s due to muscle rigidity and tremor, highlighting the phonetic markers most affected by the disease.

\vspace{-0.6cm}
\begin{table*}[htbp]
\centering
\footnotesize
\caption{Evaluation of Performance on Vowels and Extended Speech Signals}
\resizebox{0.89\textwidth}{!}{
\begin{tabular*}{\textwidth}{@{\extracolsep{\fill}}lcccc}
\toprule
\multicolumn{5}{c}{Vowels Only}
\\ 
\midrule
\textbf{Model} & \textbf{Accuracy} & 
\textbf{Precision} & \textbf{Recall} & \textbf{F1-Score} \\
\midrule 
KNN & $85.46 \pm 1.30$ & $84.95 \pm 0.87$ & $94.19 \pm 2.11$ & $78.36 \pm 3.53$ \\
        SVM & $94.37 \pm 2.41$ & $93.32 \pm 3.33$ & $93.58 \pm 6.19$ & $95.52 \pm 1.76$ \\
        RF & $92.51 \pm 3.24$ & $91.66 \pm 3.23$ & $92.34 \pm 1.77$ & $92.77 \pm 5.09$ \\
        GB & $92.38 \pm 3.08$ & $91.37 \pm 3.45$ & $96.10 \pm 1.23$ & $92.42 \pm 5.49$ \\
\midrule
\multicolumn{5}{c}{Extended Speech Signals\textsuperscript{1}}
\\
\midrule
KNN & 97.26  $\pm$  1.58 & 97.78  $\pm$  2.46 & 96.17  $\pm$  3.39 & 96.91  $\pm$  1.80 \\
SVM & 96.37  $\pm$  2.24 & 99.39  $\pm$  1.14 & 92.46  $\pm$  5.47 & 95.70  $\pm$  2.80 \\
RF & 95.42  $\pm$  2.46 & 96.63  $\pm$  3.86 & 93.23  $\pm$  3.91 & 94.82  $\pm$  2.77 \\
GB & 91.17  $\pm$  2.61 & 97.91  $\pm$  2.53 & 82.22  $\pm$  5.76 & 89.25  $\pm$  3.41 \\
\midrule
\textbf{Ours} & \textbf{99.14  $\pm$  1.60} & \textbf{100.0  $\pm$  0.0} & \textbf{98.15  $\pm$  3.46} & \textbf{99.03  $\pm$  1.81} \\
\bottomrule
\end{tabular*}
}
\label{tab:metrics}
\end{table*}

\begin{table*}[htbp]
\centering
\footnotesize
\caption{Words most indicative of PD speech with IPA transcriptions and frequency.}

\begin{tabular*}{\textwidth}{@{\extracolsep{\fill}}lcccccccccc}
\toprule
\textbf{Word} & Cocco & Ciccio & Luna & Sedeva & Neve & Mamma & Vicino & Muro & Gigi & Rete \\
\textbf{IPA} & \textipa{'kOk.ko} & \textipa{'tSItSo} & \textipa{'lu.na} & \textipa{se'de.va} & \textipa{'ne.ve} & \textipa{'mam.ma} & \textipa{vi'tSi.no} & \textipa{'mu.ro} & \textipa{'dZi.dZi} & \textipa{'re.te} \\
\textbf{Freq.} & 12 & 11 & 11 & 10 & 10 & 9 & 8 & 8 & 8 & 7\\
\bottomrule
\end{tabular*}

\label{tab:pd}
\end{table*}

\noindent\textbf{Conclusion.} In this study, we presented an explainable workflow incorporating a temporal network to analyze speech patterns in segments, allowing for a more detailed understanding of how specific phonetic units relate to Parkinson’s disease. This allowed us to identify key speech segments, such as consonants and vowels, that exhibit characteristic impairments in Parkinson’s patients. Learning the temporal dynamics of speech deterioration has the potential to offer valuable insights into the individual progression of the disease. In future studies, we will examine the model’s generalization across languages by assessing the influence of phonetic similarities and native language, while also encouraging further research into the relationship between phonation patterns and Parkinson’s progression to support personalized monitoring strategies.

\bibliographystyle{splncs04}
\bibliography{bibliography}
\end{document}